\documentclass[letterpaper]{article} 
\usepackage{aaai2026}  
\usepackage{times}  
\usepackage{helvet}  
\usepackage{courier}  
\usepackage[hyphens]{url}  
\usepackage{graphicx} 
\usepackage{natbib}  
\usepackage{caption} 
\usepackage{algorithm}
\usepackage{algorithmic}
\usepackage{booktabs}
\usepackage[inkscapelatex=false]{svg}
\usepackage{bm}
\usepackage{pifont}

\usepackage{amsmath}

\usepackage{newfloat}

\usepackage{listings}
\usepackage{multirow}
\DeclareCaptionStyle{ruled}{labelfont=normalfont,labelsep=colon,strut=off} 
\floatstyle{ruled}
\newfloat{listing}{tb}{lst}{}
\floatname{listing}{Listing}
\title{TATTOO: Training-free AesTheTic-aware Outfit recOmmendation}
\newcommand{\symfnmark}[1]{%
  \begingroup
  \renewcommand\thefootnote{\fnsymbol{footnote}}%
  \footnotemark[#1]%
  \endgroup}

\newcommand{\symfntext}[2]{%
  \begingroup
  \renewcommand\thefootnote{\fnsymbol{footnote}}%
  \footnotetext[#1]{#2}%
  \endgroup}
\author{
  Yuntian Wu\textsuperscript{\rm 1, 2}\symfnmark{2}\symfnmark{3},
  Xiaonan Hu\textsuperscript{\rm 1}\symfnmark{2},
  Ziqi Zhou\textsuperscript{\rm 3},
  Hao Lu\textsuperscript{\rm 1}\symfnmark{1}
}
\affiliations{
    \textsuperscript{\rm 1}National Key Laboratory of Multispectral Information Intelligent Processing Technology, School of Artificial Intelligence and Automation, Huazhong University of Science and Technology, Wuhan, China\\
    \textsuperscript{\rm 2}Carnegie Mellon University, Pittsburgh, United States\\
    \textsuperscript{\rm 3}School of Cyber Science and Engineering, Huazhong University of Science and Technology, Wuhan, China\\
    yuntianw@andrew.cmu.edu, \{xiaonah,zhouziqi,hlu\}@hust.edu.cn    
}

\usepackage{bibentry}

\begin{document}

\maketitle
\symfntext{2}{These authors contributed equally.}
\symfntext{3}{Part of work was done when Yuntian Wu was with Huazhong University of Science and Technology.}
\symfntext{1}{Corresponding author.}
\begin{abstract}
The global fashion e-commerce market relies significantly on intelligent and aesthetic-aware outfit-completion tools to promote sales. While previous studies have approached the problem of fashion outfit-completion and compatible-item retrieval, most of them require expensive, task-specific training on large-scale labeled data, and no effort is made to guide outfit recommendation with explicit human aesthetics. In the era of Multimodal Large Language Models (MLLMs), we show that the conventional training-based pipeline could be streamlined to a training-free paradigm, with better recommendation scores and enhanced aesthetic awareness. We achieve this with TATTOO, a Training-free AesTheTic-aware Outfit recommendation approach. It first generates a target-item description using MLLMs, followed by an aesthetic chain‑of‑thought used to distill the images into a structured aesthetic profile including color, style, occasion, season, material, and balance. By fusing the visual summary of the outfit with the textual description and aesthetics vectors using a dynamic entropy‑gated mechanism, candidate items can be represented in a shared embedding space and be ranked accordingly. 
Experiments on a real‑world evaluation set Aesthetic-$100$ show that TATTOO achieves state-of-the-art performance compared with existing training-based methods. Another standard Polyvore dataset is also used to measure the advanced zero-shot retrieval capability of our training-free method. 
\end{abstract}

\begin{figure}[!t]
    \centering
    \includegraphics[width=\linewidth]
    {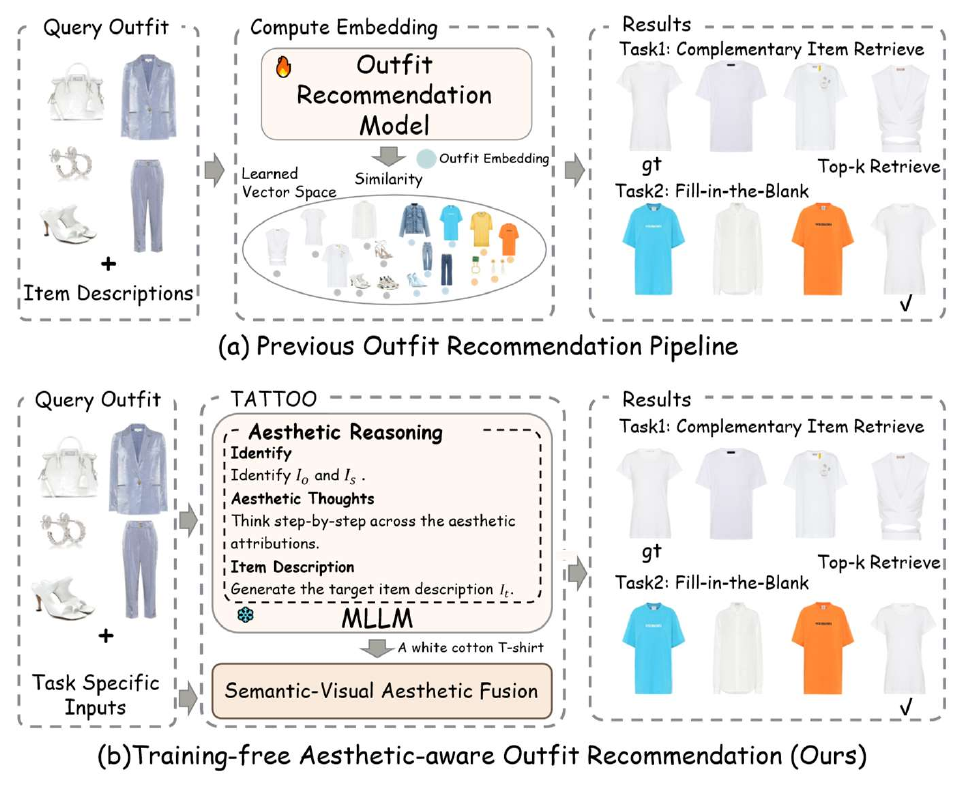}
    \caption{\textbf{Comparison between the previous training-based and our training-free outfit recommendation paradigm.} (a) Previous methods learn a latent embedding space and retrieve complementary items via similarity vector search, yielding recommendations that are accurate but opaque to human interpretation. (b) Our training-free paradigm leverages an MLLM to reason aesthetic attributes explicitly, producing both accurate and 
    interpretable results.}
    \label{fig:intro}
\end{figure}
\section{Introduction}
Fashion industry is expanding rapidly. By 2025, the global apparel market is expected to reach $\$1.84$ trillion, accounting for approximately $1.63\%$ of global GDP, with more than $100$ billion pieces of garments estimated to be produced in the same year. Faced with such a vast array of products, $82\%$ of consumers 
expect to use AI to reduce shopping 
browsing time~\cite{UniformMarket2025}. 
These statistics 
highlight the practical value of efficient outfit recommendation models. 
In outfit recommendation, Complementary Item Retrieval (CIR) and Fill-in-the-Blank (FITB) are two mainstream 
tasks for evaluating retrieval capability, where the CIR task demands completing a partial outfit by retrieving a compatible item from a large database, and the FITB requires selecting the best item from several candidates. 

Most existing work~\cite{lin2020fashion, sarkar2023outfittransformer} 
addresses CIR and FITB by learning a shared latent space in which compatibility is measured by vector distance, 
as shown in Fig~\ref{fig:intro}(a). 
This paradigm, however, has two key limitations. 
First, 
it needs \textit{time-consuming training} on large-scale, 
finely labeled datasets. Considering that fashion trends change rapidly, whether the models learned offline can catch up with the trend remains unknown. 
Second, the learned high-dimensional item embedding \textit{lacks interpretability}. Explaining a recommendation, therefore, requires additional post-hoc projections or visualizations. Although recent work~\cite{wang2024deciphering} attempts to improve interpretability by adding textual descriptions based on compatibility modeling, 
they can only explain why two items match but fails to summarize the overall aesthetics of an outfit. Furthermore, real-world styling decisions typically require aesthetic factors, including but not limited to material compatibility, style consistency, and overall balance; but none of which are explicitly considered in existing methods.



According to our preliminary study in Sec.~\ref{Sec:Ms}, we found that merely 
injecting color and material in the query descriptions can increase CIR AUC by $0.37\%$ and FITB accuracy by $5.53\%$, revealing the considerable yet under-explored application potential of \textit{aesthetic attributes} in outfit recommendation. 
This 
observation, along with two limitations mentioned above, inspires \textbf{TATTOO}, a \textit{\underline{\textbf{T}}raining-free \underline{\textbf{A}}es\underline{\textbf{t}}he\underline{\textbf{T}}ic-Aware \underline{\textbf{O}}utfit rec\underline{\textbf{O}}mmendation} approach.
As shown in Fig.~\ref{fig:intro} (b), we frame TATTOO in two stages. 
Stage \uppercase\expandafter{\romannumeral1}, \textit{Aesthetic Reasoning}, aims to address the issue of training and poor interpretability. We leverage a frozen MLLM to generate aesthetic thoughts and textual descriptions of the missing items given a partial outfit, converting its latent reasoning into a structured aesthetic profile without any gradient update, which is not afforded by prior Large Language Models (LLMs) for recommendation frameworks. 
Further in stage \uppercase\expandafter{\romannumeral2}, we propose a \textit{Semantic–visual Aesthetic Fusion} mechanism to fuse fine-grained aesthetic attributes to construct a query vector for zero-shot multimodal retrieval. 
This process simulates the thinking process of professional stylists when evaluating aesthetic elements such as color and material, simultaneously improving the model's aesthetic awareness, recommendation accuracy, and output interpretability, without the need for additional 
training. 

Experiments demonstrate that TATTOO achieves state-of-the-art performance on the aesthetic ability evaluation benchmark Aesthetic-$100$~\cite{zou2022good} and reports competitive FITB and CIR performance on the Polyvore dataset~\cite{vasileva2018learning}. A series of ablation studies are also conducted to 
validate the aesthetic ability of TATTOO and its effectiveness in complementary item retrieval.

Our main contributions can be summarized as follows:
\begin{itemize}
    \item We rethink the outfit recommendation task from the perspective of aesthetics, discovering the importance of aesthetic attributes through experiments.
    \item TATTOO: one of the first training-free outfit recommendation methods that unleashes the reasoning ability of MLLMs in the aesthetic domain and achieves effective yet human-interpretable outfit recommendations.
\end{itemize}

\section{Related Work}
\subsection{Outfit Recommendation}
Compatibility Prediction (CP) and Complementary Item Retrieval (CIR) are the two mainstream tasks in outfit recommendation~\cite{sarkar2023outfittransformer}.
CP aims to judge whether a complete outfit is harmonious. 
Existing methods typically address this by metric learning, mapping individual fashion items into a shared latent embedding space~\cite{han2017learning,li2020hierarchical,balim2023diagnosing,jing2023category}. 
CIR, on the other hand, treats a partial outfit as a query, retrieving the most complementary item from a large database. Compared to pair-wise modeling in CP, CIR highlights end-to-end outfit modeling~\cite{sarkar2023outfittransformer}. 
Notably, these two tasks are not entirely independent. Metric learning-based CP models can also use the same learned embeddings for retrieval~\cite{vasileva2018learning,tan2019learning,lin2020fashion}.
With the rise of generative AI, methods that leverage diffusion models to synthesize new fashion items 
have emerged~\cite{xu2024diffusion}, providing inspirations for fashion stylists. 
Additional work further explores the explanations for outfit recommendation. \citep{lin2019explainable} generates text explanation for top and bottom item matching, \cite{kaicheng2021modeling} introduces learnable attributes extraction to CP, and~\cite{wang2024deciphering} constructs an attribute extraction pipeline along with a pair-wise item recommendation explanation dataset. 

To the best of our knowledge, all these methods require training on specific fashion datasets. Whether competitive accuracy and rich interpretability can be achieved without any task-specific training remains an open question, especially in the era of MLLMs.

\subsection{LLM for Recommendation}
Existing works on integrating LLMs with recommendation systems can 
be categorized into several 
streams~\cite{wu2024survey}.
One major stream is embedding-based approaches, which 
considers LLMs as feature encoders to extract latent representations used for traditional recommendation models~\cite{cui2022m6,liu2024once}.
Another prominent direction is token-based methods, where LLMs are promoted to generate discrete tokens that serve as semantic-aware representations for downstream recommendation methods~\cite{zhai2023knowledge, TokenRec}.
Going a step further, end-to-end solutions treat LLMs as the recommendation system directly, removing the need for an 
intermediate model~\cite{geng2022recommendation, hou2024large}.
Recently, the landscape has expanded to MLLMs. Exemplary work includes fine-tuning MLLMs for sequential recommendation tasks~\cite{wu2025aligning, ye2025harnessing}, and integrating MLLMs with pretrained Vision-Language Models (VLMs) such as CLIP~\cite{radford2021learning} for multimodal retrieval~\cite{karthik2024visionbylanguage, tang2025reason}.

An emerging line of research has begun to incorporate LLMs or VLMs into fashion recommendation~\cite{zhaoetal2024unifashion,liu2024sequential, gao2024fashiongpt, song2024fashiongpt, pang2025fashionm3}, targeting different fashion scenarios including virtual try-on, personalized fashion recommendation, and fashion understanding. These methods, however, invariably require task-specific fine-tuning. In contrast, our method leverages a frozen MLLM in a fully training-free manner and achieves competitive performance without any task-level gradient updates, highlighting the methodological novelty of our approach.
\begin{figure*}[!t]
    \centering
    \includegraphics[width=\linewidth]{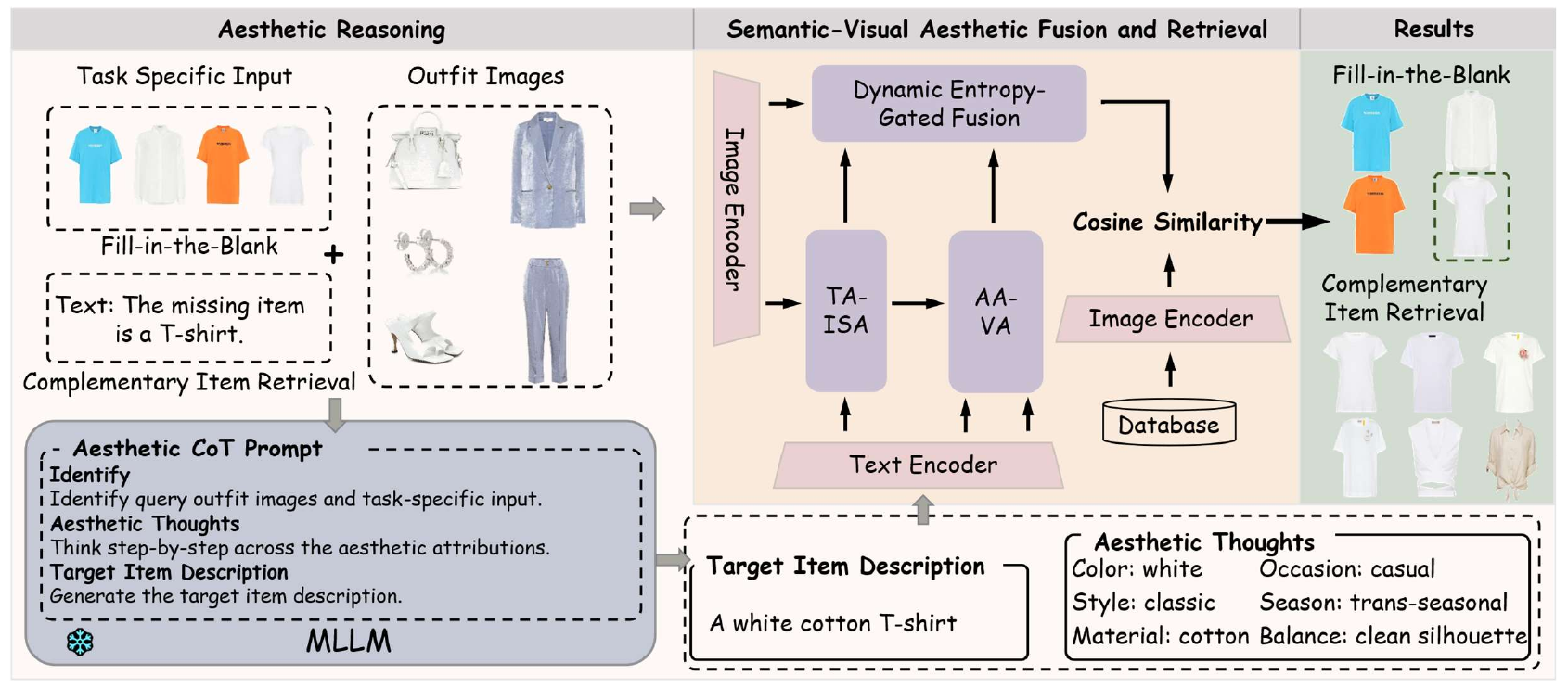}
    \caption{\textbf{Overview of our technical pipeline.} We frame TATTOO into two stages. In stage I, we utilize an MLLM to reason the target missing item taking aesthetic attributes and interpretability into account. In stage II, we design a Target‑Aware Image Saliency Aggregator (TA-ISA), an Adaptive Aesthetics Vector Aggregator (AA-VA), and a Dynamic Entropy-Gated Fusion (DE-GF) to construct the query vector based on both target item description and aesthetic attributes.  }
    \label{fig:me.pipeline}
\end{figure*}

\section{Observation and Motivation}\label{Sec:Ms}

State-of-the-art outfit recommendation methods generally rely on both image and text modalities to represent a fashion item. An ideal textual description should 
align with visual cues with semantic ones. 
However, in most public outfit recommendation benchmarks, textual 
descriptions are crawled from web pages or user posts directly, presenting noise 
such as emotive phrases, marketing slogans, and meaningless symbols like emojis, which weakens vision-language alignment~\cite{zou2022good}. 
We argue that \textit{such coarse labels impede the model to learn and perceive intra-outfit compatibility
}.
Consequently, we are motivated to devise \textit{fine-grained and semantically accurate} item descriptions. 

To validate this hypothesis, we reconstruct the textual descriptions in Polyvore~\cite{vasileva2018learning}, a mainstream CP and FITB dataset,
and conduct preliminary experiments with three textual baselines: 
i) \textit{Vision only}, where the item descriptions are removed entirely; 
ii) \textit{Original captioning}, in which the raw, noisy item descriptions are retained; 
and iii) \textit{Fine-grained captioning}, where we inject two attributes of color and material into each original item description.
We adapt the 
OutfitTransformer~\cite{sarkar2023outfittransformer}, replace its image encoder and text encoder with FashionCLIP~\cite{chia2022contrastive}, and train the models for CP and FITB, 
respectively. 
\begin{table}[!t]
\centering
\begin{tabular}{@{}lcc@{}}
\toprule
Baselines & {AUC (\%)$\uparrow$} & {ACC (\%)$\uparrow$} \\
\midrule
Vision only      & 88.79 & 62.40 \\
Original captioning     & 95.01 & 67.43 \\
Fine-grained captioning & \textbf{95.37} & \textbf{71.16} \\
\bottomrule
\end{tabular}
\caption{\textbf{Comparison between different textual description settings.} Best results are in \textbf{boldface}.}
\label{ms.pre}
\end{table}
According to Table~\ref{ms.pre}, we find that modifying the initial item descriptions with color and material attributes yields improved AUC and ACC on both CP and FITB. This indicates that even small amounts of fine-grained item descriptions can improve the accuracy of recommendations.

Indeed, factors such as color and material are highly relevant to the visual aesthetics of a fashion item, providing semantic cues that align with consumer preferences, thus helping the model recommend more contextually appropriate items. 
This observation subsequently raises an assumption: \textit{whether incorporating additional 
fine-grained, aesthetic features 
can benefit outfit recommendation?} 
To validate this hypothesis, we conduct a systematic review of existing outfit recommendation methods and evaluation protocols, and select candidate attributes based on the coverage of public datasets and semantic diversity, the latter of which refers to the granularity of an attribute in the fashion context (i.e., attribute color contributes an extremely rich vocabulary in fashion scenarios). 
We adapt the six attributes summarized in~\cite{zou2022good}: \textit{color}, \textit{style}, \textit{occasion}, \textit{season}, \textit{material}, and \textit{balance}, which includes the two attributes that are validated in our preliminary experiments. In this work, we collectively refer to these six features as \textit{aesthetic attributes}. For details of attribute selection and theoretical insight, please refer to our Supplementary. 


\section{Proposed Method}
\begin{figure*}[!t]
    \centering
    \includegraphics[width=\linewidth]
    {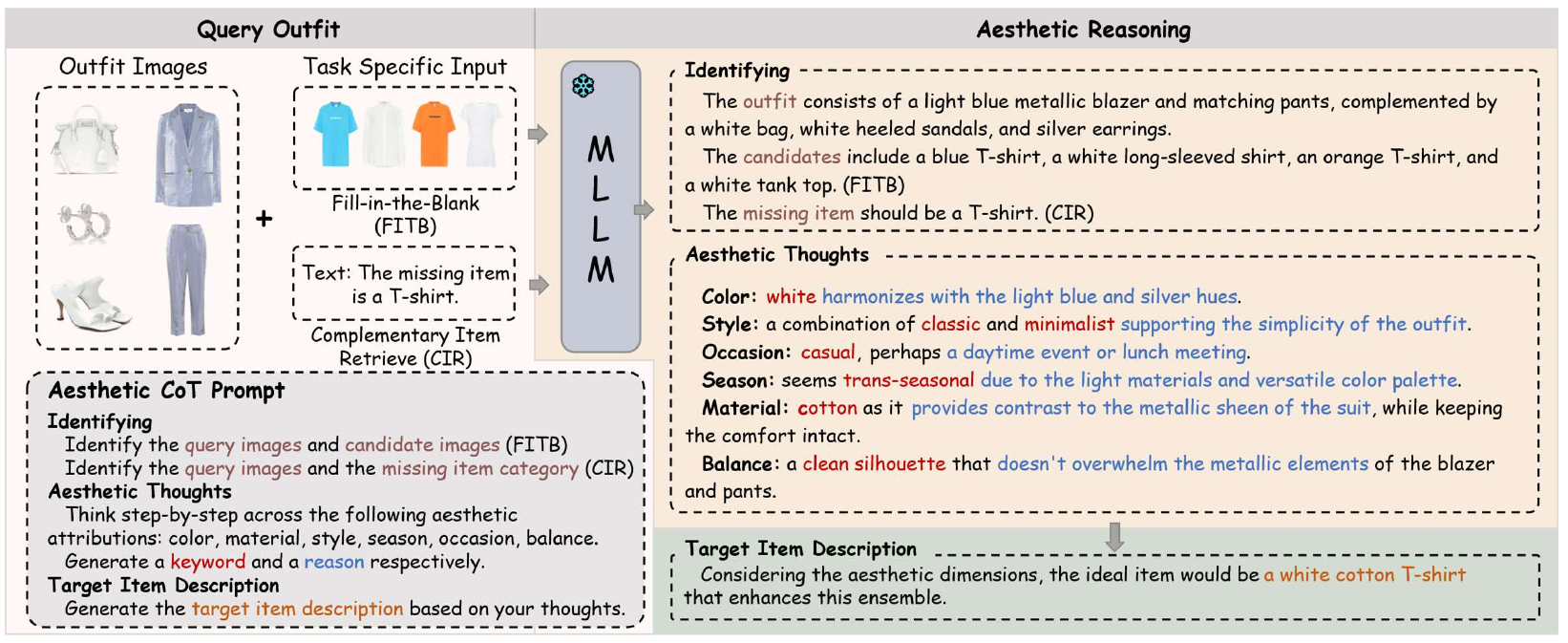}
    \caption{\textbf{Aesthetic reasoning process.} We design an aesthetic CoT prompt, instructing the MLLM to generate an image description $T_{t}$ and aesthetic thoughts $T_a$ of the missing item, which are later used to construct a query vector for retrieval.}
    \label{fig:me.prompt}
\end{figure*}

\subsection{Problem Formulation}
For 
outfit recommendation, 
we define a user query $Q$ as $(\bm I_o,\bm I_{s})$, where $\bm I_{o} =\{\bm{I}^{(i)}_{o}\}_{i=1}^{n}$ denotes images of $n$ items forming a partial outfit, and $\bm I_{s}$ denotes the task-specific input: candidate item images in FITB and textual category of the missing item.
Our goal is to retrieve the missing item image $\bm{\hat{I}}$ from the database $\mathcal{D}$. We frame this task as a two-stage multimodal reason-and-retrieve 
problem. 
As shown in Fig.~\ref{fig:me.pipeline}, we first utilize an MLLM $\Psi_{M}$ to reason the image description $T_{t}$ of the target missing item in \textit{aesthetic reasoning}. To make this stage 
interpretable for humans, we incorporate an \textit{aesthetic CoT prompt} $p_a$ to instruct a rational reasoning process 
w.r.t. the following aesthetic attributes: color, style, occasion, material, season, and balance. 
By symbolizing the generated aesthetic thoughts 
as $T_{a} = \{T_{a}^{(i)} | i\in({\tt col, sty, occ, sea, mat,bal})\}$, 
the reasoning process could be jointly represented by
\begin{equation}\label{equ:pipeline}
    T_{t}, T_{a} = \Psi_{M}(p_a \circ\bm I_{o}\circ\bm I_{s})\,,
\end{equation}
where $\circ$ denotes sequence prompt composition.
We then use the pretrained image encoder $\Psi_{I}$ and text encoder $\Psi_{T}$ of FashionCLIP to encode the query $Q$ along with the generated textual $\bm T_{a}$ and $ \bm T_{t}$. After that, we incorporate a \textit{semantic-visual aesthetic fusion} ($\tt {SVAF}$) stage to construct a query vector 
$\bm{q}=$ $\tt{SVAF}$$(\Psi_{T}(\bm I_{o}),\Psi_{T/I}(\bm I_{s}), \Psi_{T}(T_{t}), \Psi_{T}(T_{a}))$, 
which is finally passed for item retrieval 
by 
\begin{equation}
\bm{\hat{I}} = \underset{d_I \in \mathcal{D}}{\arg\max} \frac{\bm q^\mathsf{T} \cdot \Psi_I(d_I)}{\|\bm q^\mathsf{T}\cdot\Psi_I(d_I)\|_2}\,,
\label{eq:retrieve}
\end{equation}
Considering the composition of the query vector $\bm q$, Eq.~\ref{eq:retrieve} actually achieves an aesthetic-aware and multimodal retrieval.
\subsection{Aesthetic Reasoning}
Generating the target item description $T_t$ and aesthetic attributes $T_a$ requires the MLLM to understand the relationships within the observed items and between the observed and the missing one. 
These relationships include but not limited to aesthetic thoughts, functional roles (top, bottom, outerwear), and contribution of each item to overall visual balance.
To recommend a compatible outfit, we propose the \textit{Aesthetic CoT prompt} $p_a$, which guides the model to reason step-by-step across aesthetic attributes and facilitates an aesthetically reasonable target item description $T_t$. As shown in Fig.~\ref{fig:me.prompt}, $p_a$ instructs the MLLM following three progressive reasoning steps:  
i) the \textit{Identifying} step highlights the composition of the query questions. In this step, we prompt the MLLM to recognize and generate the roles of $\bm I_{o}$ and $\bm I_{s}$ in $T_o$, interpreting their visual, semantic, and interrelationship; 
ii) \textit{Aesthetic Thoughts} then guide the MLLM to reason across the six aesthetic attributes to select the most compatible item. The reasoning results, aesthetic thoughts $T_a$, have two fields for each attribute: a \textit{keyword} briefly describing the attribute, and a \textit{reason} explaining why this attribute fits the query outfit; 
iii) \textit{Target Item Description}, where the MLLM synthesizes $T_o$ and $T_a$ into a detailed image description of the missing item. This description, $T_t$, grounded in aesthetic reasoning, serves as a precise and coherent guide for retrieving the most suitable complementary item for the outfit.
\begin{table*}[!t]
\setlength{\tabcolsep}{2pt} 
\centering
\begin{tabular}{@{}lcccccccccc@{}}
\toprule
\multirow{2}{*}{Method} & \multirow{2}{*}{Training‑free} & \multicolumn{2}{c}{LATs} & \multicolumn{7}{c}{AATs} \\
\cmidrule(lr){3-4} \cmidrule(lr){5-11}
& &\multicolumn{1}{c}{Hard} & \multicolumn{1}{c}{Soft} & Color & Style & Occasion & Season & Material & Balance & Total \\
\midrule
Bi-LSTMs~\cite{han2017learning} & \ding{55}  & 36\% & 30.82\% & 0.30 & 0.34 & 0.40 & 0.50 & 0.42 & 0.11 & 35\% \\
FHN~\cite{lu2019learning} & \ding{55}  & 54\% & 41.62\% & 0.50 & 0.22 & 0.60 & 0.42 & 0.50 & 0.33 & 40\% \\
SCE-Net~\cite{tan2019learning} & \ding{55}  & 72\% & 54.63\% & 0.75 & 0.28 & 0.33 & 0.33 & 0.50 & 0.33 & 42\% \\
CSN~\cite{vasileva2018learning} & \ding{55}  & 73\% & 56.17\% & 0.85 & 0.50 & 0.53 & 0.58 & 0.50 & 0.56 & 59\% \\
OutfitTransformer~\cite{sarkar2023outfittransformer} & \ding{55}  & 60\% & 46.44\% & 0.75 & 0.41 & 0.60 & 0.25 & 0.42 & 0.22 & 47\% \\
OutfitTransformer (F-CLIP) & \ding{55}  & 75\% & 56.49\% & 0.80 & \textbf{0.68} & 0.86 & \textbf{0.75} & 0.50 & \textbf{0.67} & 72\% \\
\textbf{TATTOO} (Ours) & \ding{51}  & \textbf{79}\% & \textbf{58.80}\% & \textbf{0.89} & 0.59 & \textbf{0.87} & 0.67 & \textbf{1.0} & 0.56 & \textbf{75}\% \\
\hline
\end{tabular}
\caption{\textbf{Comparison on A100 benchmark.} Our method outperforms all training methods on both the liberalism aesthetic test and the academicism aesthetic test, indicating a robust and generalizable aesthetic ability. Best performances are in \textbf{boldface}. }
\label{tab:A100}
\end{table*}
\subsection{Semantic–Visual Aesthetic Fusion}

Albeit prompted to derive $T_t$ based on $T_a$, the MLLM may still overlook certain attribute dimensions. 
To ensure a target item description informed and aesthetic-attribute-aware retrieval, 
we propose a Semantic-Visual Aesthetic Fusion mechanism to construct the query vector $\bm q$.

\subsubsection{Target‑Aware Image Saliency Aggregator (TA‑ISA).} In this step, we quantify how well the query images $\bm I_{o}=\{\bm{I}_{o}^{(i)}\}_{i=1}^{n}$ align with $T_t$. Given an embedding of one query image $\bm v_{o}^{(k)}={\Psi_{I}(\bm I_{o}^{(k)})}/{||\Psi_{I}(\bm I_{o}^{(k)})||_{2}}$ and the embedding  of the textual target item description $\bm v_{t}={\Psi_{T}(T_{t})}/{||\Psi_{T}(T_{t})||_{2}}$, TA‑ISA assigns soft saliency weights to each query item image by
\begin{equation}
w_{o}^{(k)}=\frac{\exp\!\bigl(\tau^{-1}\,\bm{v}_{t}^\mathsf{T}\bm v_{o}^{(k)}\bigr)}
            {\textstyle\sum_{j=1}^{n}\exp\!\bigl(\tau^{-1}\,
                  \bm{v}_{t}^\mathsf{T}\bm v_{o}^{(j)}\bigr)}\,,k=1,2,...,n\,,
\label{eq:saliency_w}
\end{equation}
where $\tau$ is the temperature. Eq.~\ref{eq:saliency_w} retains only the visual components that are relevant for describing the missing item, thus filtering noisy garments. This soft saliency weight is then applied to mitigate the impact of visually prominent but task-irrelevant factors by
\begin{equation}
\bm v_{I}=\frac{\sum\nolimits_{k=1}^{n}w_{o}^{(k)}\bm v_{o}^{(k)}}{\Bigl\|
           \sum\nolimits_{k=1}^{n}w_{o}^{(k)}\bm v_{o}^{(k)}\Bigr\|_{2}}\,.
\label{eq:vI}
\end{equation}
where the $\bm v_I$ denotes the \textit{target‑aware visual embedding}. In effect, $\bm v_I$ condenses the visual context into a fixed‑length vector that emphasises features most predictive of the missing slot while suppressing distractors, thereby providing a comparatively clean query signal for downstream tasks.  

\subsubsection{Adaptive Aesthetics Vector Aggregator (AA‑VA).} Given the aesthetic thoughts $T_a=\{T_a^{(i)}|i\in(\tt{col,sty,occ,sea,mat,bal})\}$, we need to adaptively consider which and to what extent one aesthetic attribute needs to be compensated for the $T_t$. 
We first quantify the degree of compensation by
\begin{equation}
s_{d}^{(i)}= \frac{\bigl({\Psi_{T}(T_{a}^{(i)})^{\mathsf{T}}}\bm v_{t}+{\Psi_{T}(T_{a}^{(i)})^{\mathsf{T}}}\bm v_{I}\bigr)}{2}\,.
\label{eq:sd}
\end{equation}
Consequently, we use this dynamic correlation to encode all aesthetic attributes as
\begin{equation}
\bm v_{AES}= \frac{\sum\nolimits_{i=1}^{m} exp(s_{d}^{(i)}) \Psi_{T}(T_{a}^{(i)})}{\Bigl\|\sum\nolimits_{i=1}^{m}exp(s_{d}^{(i)})\Psi_{T}(T_{a}^{(i)}) \Bigr\|_{2}}\,.
\label{eq:Vaes}
\end{equation}
This integration method dynamically enhances $T_a^{(i)}$ that are not considered in $T_t$, thereby encoding the aesthetic knowledge adaptively.

\subsubsection{Dynamic Entropy‑Gated Aggregation (DE‑GF).} 
To meet diverse retrieval needs, we dynamically merge $\bm v_I$, $\bm v_t$, and $\bm v_{AES}$ into a single query vector while keeping it in the original vision–language embedding space.
For each $\bm v\in\mathcal V=\{\bm v_I, \bm v_t, \bm v_{AES}\}$, we compute a candidate‑level similarity distribution $p(\mathbf v)=$ $\tt {softmax}$$(\Psi_{T/I}(I_s),\mathbf v)$, and measure its entropy $H(\mathbf v)=-\sum_{i}p_i\log p_i$. A lower entropy indicates that the cue already pinpoints a compact subset of promising matches; therefore, a gating weight $g_\mathbf v=\exp\bigl(-H(\mathbf v)\bigr)$ is assigned and normalized across cues. The final query representation can be denoted as
\begin{equation}
\bm q= \frac{\sum\nolimits_{\bm v\in\mathcal V} g_{v}\,\bm v }{\Bigl\|
       \sum\nolimits_{\bm v\in\mathcal V} g_{v}\,\bm v
       \Bigr\|_{2}}\,.
\label{eq:q_final}
\end{equation}
Since the gates are computed on‑the‑fly from the candidate set, DE–GF dynamically balances visual grounding, textual semantics, and aesthetic guidance according to the ambiguity of each retrieval episode.

\begin{table*}[!t]
\centering
\setlength{\tabcolsep}{4pt}   
\begin{tabular}{lccccccccc}
\toprule
\multirow{2}{*}{Method} &
\multirow{2}{*}{Training-free} &
\multicolumn{4}{c}{Polyvore-D} &
\multicolumn{4}{c}{Polyvore} \\
\cmidrule(lr){3-6} \cmidrule(lr){7-10}
& & ACC (\%) & R@10 & R@30 & R@50 & ACC (\%) & R@10 & R@30 & R@50 \\
\midrule
CSN~\cite{vasileva2018learning} & \ding{55}          & 55.65 & 3.66 & 8.26 & 11.98 & 57.83 & 3.50 & 8.56 & 12.66 \\
SCE-Net~\cite{tan2019learning}  & \ding{55}            & 53.67 & 4.41 & 9.85 & 13.87 & 59.07 & 5.10 & 11.20 & 15.93 \\
CSA-Net~\cite{lin2020fashion}   & \ding{55}          & 59.26 & 5.93 & \textbf{12.31} & \textbf{17.85} & 63.73 & 8.27 & 15.67 & 20.91 \\
OutfitTransformer~\cite{sarkar2023outfittransformer} & \ding{55}   & 59.48 & \textbf{6.53} & 12.12 & 16.64 & \textbf{67.10}& \textbf{9.58} & \textbf{17.96} & \textbf{21.98} \\
\textbf{TATTOO} (Ours) & \ding{51}       & \textbf{65.74} & 6.16 & 10.99 & 15.70 & 65.98 & 5.39 & 10.11 & 14.08 \\
\bottomrule
\end{tabular}
\caption{\textbf{Comparison on the Polyvore-D and Polyvore}. Our training-free method achieves SOTA on the Polyvore-D FITB task, and achieves comparable performance on the Polyvore-D CIR, Polyvore FITB and CIR tasks. Best results are in \textbf{boldface}. }
\label{tab:polyvore}
\end{table*}

\begin{figure*}[!t]
    \centering
    \includegraphics[width=0.9\linewidth]
    {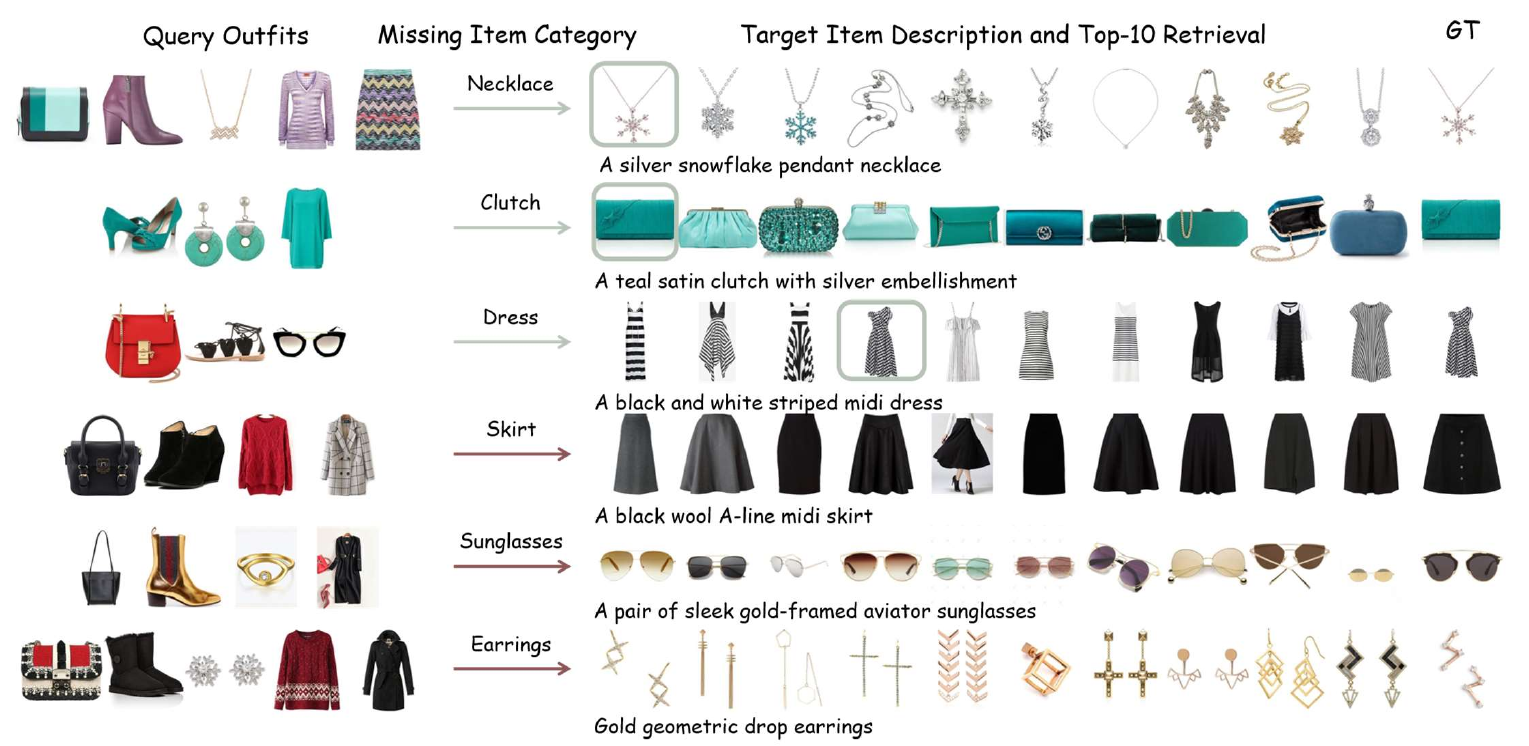}
    \caption{\textbf{Visualization of CIR on the Polyvore dataset.} Green arrows symbolize correct retrievals, while red arrows indicate failure cases. Note that items in failure cases may not necessarily be incompatible with the query outfits.}
    \label{fig:exp.vis}
\end{figure*}

\section{Results and Discussion}

\subsection{Experimental Setup}
\subsubsection{Datasets and Protocol.} We utilize the Aesthetic-$100$ protocol (A100)~\citep{zou2022good} and the Polyvore dataset~\cite{vasileva2018learning} for evaluation. A100 is a question-based benchmark comprising two FITB tests, namely the Liberalism Aesthetic Test (LAT) and the Academicism Aesthetic Test (AAT), focusing on crowd-consensus aesthetics and expert-driven fashion principles, respectively. 
Polyvore is one of the mainstream benchmarks for outfit recommendation, supporting both the CP and FITB tasks. It provides the disjoint (Polyvore-D) and nondisjoint (Polyvore) splits. The nondisjoint set consists of $53,306$ outfits for training and $10,000$ outfits for testing, where items may appear across the training, validation, and testing subsets. The disjoint split contains $16,995$ training outfits and $15,154$ test outfits,  where no item overlaps across training, validation, and testing. We further adapt the Polyvore for the CIR task following~\cite{lin2020fashion}.

\subsubsection{Baselines.} We compare with the following baselines: Bi-LSTM~\cite{han2017learning}, CSN~\cite{vasileva2018learning}, FHN~\cite{lu2019learning}, SCE-Net~\cite{tan2019learning}, CSA-Net~\cite{lin2020fashion}, and OutfitTransformer~\cite{sarkar2023outfittransformer}. We highlight the OutfitTranformer. Among post-2023 work, OutfitTransformer is the only approach that addresses CIR and FITB tasks, and reports results on the Polyvore under the identical evaluation protocol we adapt, making it the most relevant SOTA baseline. Since this method has no publicly available official implementation, we re-implement it based on the algorithm details reported in the original paper. We further introduce an improved variant with its original image encoder and text encoder replaced by FashionCLIP. In the rest of the paper, we use \textit{F-CLIP} to indicate the improved variant. For details of other baselines, please refer to our Supplementary.
\subsubsection{Evaluation Metrics.} To evaluate the model's aesthetic ability, we utilize LATs (hard), mLATs (soft), and AATs introduced by A100~\cite{zou2022good}. Further, we compute FITB ACC and Recall@top-K (R@K) on the Polyvore dataset to evaluate the recommendation accuracy and retrieval capability. Details can refer to the Supplementary.

\subsubsection{Implementation Details} The default MLLM used in our method is GPT-4o~\cite{OpenAI2024GPT4o}. We also perform ablation studies with open-source MLLMs including LLaMA~\cite{MetaAI2025Llama4} and Qwen~\cite{bai2023qwenvl}. We interface with all MLLMs via the official OpenAI SDK. Specifically, we set the temperature to $0$, and all other parameters use the default values. We set $\tau=0.01$ in TA-ISA.
\begin{table}[!t]
\centering
\setlength{\tabcolsep}{3pt}
\begin{tabular}{@{}lccccccc@{}}
\toprule
MLLM  & Ide. & $\tt SVAF$ & Aes. & LATs & mLATs & AATs \\ 
\midrule
\multirow{3}{*}{GPT-4o} 
 & \ding{55} &  &  & 78\% & 59.02\% & 72\% \\  
 &  & \ding{55} &  & 77\% & 58.69\% & 72\% \\ 
 &  & \ding{55} & \ding{55} & 74\% & \textbf{59.54\%} & 66\% \\
 & \ding{51} & \ding{51} & \ding{51} & \textbf{79\% }& 58.80\% & \textbf{75\%} \\  \midrule
\multirow{3}{*}{LLaMA-16e} 
 & \ding{55} &  &  & 51\% & 39.83\% & 52\% \\  
 &  & \ding{55} &  & 55\% & 42.78\% & 54\% \\  
 &  & \ding{55} & \ding{55} & 49\% & 41.37\% & 50\% \\ 
 & \ding{51} & \ding{51} & \ding{51} & \textbf{62\%} & \textbf{47.58\%} & \textbf{59\%} \\ \midrule
\multirow{3}{*}{LLaMA-128e}  
 & \ding{55} &  &  & 53\% & 43.37\% & 49\% \\  
 &  & \ding{55} &  & 74\% & 55.75\% & 53\% \\ 
 &  & \ding{55} & \ding{55} & 70\% & 54.69\% & 47\% \\ 
 & \ding{51} & \ding{51} & \ding{51} & \textbf{75\%} & \textbf{56.42\%} & \textbf{64\%} \\\midrule
\multirow{3}{*}{Qwen-v1} 
 & \ding{55} &  &  & 61\% & 46.85\% & 55\% \\  
 &  & \ding{55} &  & 76\% & 56.16\% & 64\% \\ 
 &  & \ding{55} & \ding{55} & 73\% & 53.65\% & 61\% \\ 
 & \ding{51} & \ding{51} & \ding{51} & \textbf{77\%} & \textbf{56.95\%} & \textbf{66\%} \\ 
\bottomrule
\end{tabular}
  \caption{\textbf{Ablation study on the core components and different based MLLMs}. \textit{Ide.} is short for \textit{Identifying}, and \textit{Aes.} is short for \textit{Aesthetic Thoughts}. \ding{55} denotes disabling the corresponding step. Best results are in \textbf{boldface}.}
  \label{tab:ablation}
\end{table}

\subsection{Main Results}
Our main results are presented in Table~\ref{tab:A100},~\ref{tab:polyvore} and~\ref{tab:ablation}. Visualizations of complementary item retrieval are shown in Fig.~\ref{fig:exp.vis}.

In Table~\ref{tab:A100}, we compare 
the aesthetic ability between our method and other baselines. The results show that our training-free solution significantly outperforms the baselines under both mixed public aesthetic and expert aesthetic scenarios, with $4\%$ improvement in LATs (hard), $2.31\%$ in mLATs (soft), and $3\%$ in AATs compared to the best-performing baseline. This improvement indicates that our method is better at capturing the overall aesthetic characteristic of the outfit. 
By further probing the six fine-grained aesthetic attributes, TATTOO ranks the first in the color, material, and occasion, which is consistent with the tendency observed in our preliminary experiments. These results validate the necessity of aesthetic attribute modeling.

We further provide FITB and CIR results on the Polyvore in Table~\ref{tab:polyvore}. 
We cite CSN, SCE-Net, and CSA-Net directly from the original paper.  
Notably, without any task-specific training, our method still achieves SOTA performance on the disjoint split w.r.t. the FITB task and comparable performance on the nondisjoint split of the same task.
In terms of R@K, our method achieves comparable performance to OutfitTransformer on the disjoint split, and to SCE-Net on the nondisjoint split. 
This limitation is closely related to Polyvore's annotation method of CIR: for a partial outfit, 
only one single correct label is retained, 
leaving all other compatible combinations negative, while in real-world styling scenarios, one query outfit often corresponds to multiple compatible and coordinated alternative items. This point is also emphasized by~\citet{lin2020fashion}. 

We believe that the neglect of multi-solution characteristics underestimates the performance of our method. To substantiate this claim, we visualize the CIR results of our method. As shown in Fig.~\ref{fig:exp.vis}, in cases flagged as failures, our method actually retrieves items that are compatible with the query outfit: the skirts in the fourth row are virtually indistinguishable from the ground truth in color (clack), material (wool), and balance (A-line, midi); the sunglasses in the fifth row closely match the ground-truth on color (gold) and contour (sleek); and the earrings in the sixth row also adhere to the same style (geometric) aesthetic. It is worth noting that, unlike all baselines that rely on training and may implicitly memorise the uniquely annotated positives, our method is \textit{training-free} and has never seen the Polyvore dataset. This characteristic makes our method more susceptible to penalties under the current CIR setting. This effect could quantitatively be evidenced by the performance gap between the disjoint and nondisjoint splits: all baselines record markedly higher scores on the nondisjoint split, whereas our method stays virtually unchanged. 
These observations indicate that disregarding multiple compatible items of the current CIR evaluation framework is the primary reason for our method fall short of the reported SOTA. Rather than R@K, our method delivers \textit{fine-grained and interpretable recommendations} encompassing color, material, season, style, and balance across all examples. This capability, absent in prior work, constitutes a more practical contribution. For additional visualizations, please refer to our Supplementary.



\subsection{Ablation Study}

In this section, we examine the contributions of different parts of our method, along with the impact of different MLLMs. We select LLaMA-4-Scout 17B/16E (LLaMA-16e), LLaMA-4-Maverick 17B/128E (LLaMA-128e), and Qwen-VL-Max (Qwen-vl) as different base MLLMs, and conduct ablation experiments on all models, with one key component removed at a time. 
Results are shown in Table~\ref{tab:ablation}. 
\subsubsection{Impact on Core Components.} In this section, we verify the effect of aesthetic reasoning and $\tt SVAF$. 
For all variants based on different MLLMs, removing either the \textit{Identifying} (Ide.) or $\tt SVAF$ individually results in measurable performance drops. Notably, removing \textit{Aesthetic Thoughts} (Aes.) together with $\tt SVAF$ always hurts more than disabling $\tt SVAF$ alone, which aligns with the design that without the aesthetic thoughts $T_a$, $\tt SVAF$ cannot be executed, resulting in losses of both semantic prior and perspective filtering. 
Additionally, the magnitude of degradation is correlated with the strength of MLLMs. Removing the $\tt SVAF$ and aesthetic thoughts causes drops on LATs and AATs by $5\%$, and $9\%$, respectively, for GPT-4o-based variants,  whereas for LLaMA-16e-based variants, the LATs, mLATs, and AATs drop by $13\%$, $ 6.21\%$, and $9\%$. This trend generally shows that the stronger the MLLM, the smaller the performance decline.

\subsubsection{Impact on Different MLLMs.} In this section, we explore the impact of different MLLMs. 
The GPT-4o-based variant achieves the best LATs, mLATs, and AATs. For open-source MLLMs, LLaMA-128e-based and Qwen-vl-based variants achieve comparable crow-consensus metrics LATs and mLATs to the GPT-4o-based variant, and the Qwen-v1-based variant even surpasses the supervised baseline OutfitTransformer (F-clip.).
However, for the expert-driven metric AATs, the three open-source-based variants are $9-16\%$ lower than GPT-4o-based variants, indicating that professional-level aesthetic judgments still require more powerful closed-source MLLM semantic prior knowledge.
Notably, the relative changes produced by any ablation are highly consistent, validating the transferability and robustness of this method across different MLLMs.

\section{Conclusions}
In this paper, we propose a training-free outfit recommendation method that leverages MLLMs to process aesthetic attributes in the fashion domain. By perceiving aesthetic attribution of query outfit, TATTOO presents robust aesthetic reasoning and retrieval capabilities, significantly outperforming prior work on aesthetic ability evaluation, and achieving FITB accuracy and R@K comparable to the best fully-trained SOTA models. This work advances the application of aesthetic attribution on outfit recommendation and has broad implications for MLLMs in the fashion industry.

\paragraph{Limitations.}  The notion of aesthetic attributes is by no means limited to the six dimensions we studied in this paper. Further, the current pipline's best results depend on closed-source GPT-4o. The discovery of more granular, cross-cultural, or context-specific aesthetic attributes, along with the attempt for avoiding the high cost of closed-source API, remain dependent on ongoing exploration in future research.

\section{Acknowledgment}
This work was supported by the National Natural Science Foundation of China under Grant No.~62576146.

\bibliography{aaai2026}

\end{document}